\documentclass[letterpaper]{article} 
\usepackage{aaai2026}  
\usepackage{times}  
\usepackage{helvet}  
\usepackage{courier}  
\usepackage[hyphens]{url}  
\usepackage{graphicx} 
\usepackage{natbib}  
\usepackage{caption} 
\usepackage{algorithm}
\usepackage{algorithmic}
\usepackage{pifont}
\usepackage{booktabs}
\usepackage{multirow}
\usepackage{amsmath}
\usepackage{amssymb}
\usepackage{amsfonts}
\usepackage{newfloat}
\usepackage{listings}
\DeclareCaptionStyle{ruled}{labelfont=normalfont,labelsep=colon,strut=off} 
\floatstyle{ruled}
\newfloat{listing}{tb}{lst}{}
\floatname{listing}{Listing}
\title{MSPCaps: A Multi-Scale Patchify Capsule Network with Cross-Agreement Routing for Visual Recognition}
\author{
    Yudong Hu\textsuperscript{\rm 1}\thanks{Work done during an internship at FNii-Shenzhen, CUHK-Shenzhen.},
    Yueju Han\textsuperscript{\rm 2},
    Rui Sun\textsuperscript{\rm 1,\rm 3}\thanks{Corresponding authors: Jinke Ren (jinkeren@cuhk.edu.cn); Rui Sun (ruisun@link.cuhk.edu.cn)},
    Jinke Ren\textsuperscript{\rm 1,\rm 3}\footnotemark[2]
}
\affiliations{
    \textsuperscript{\rm 1}Shenzhen Future Network of Intelligence Institute, The Chinese University of Hong Kong (Shenzhen), Shenzhen, China\\
    
    \textsuperscript{\rm 2}School of Future Technology, South China University of Technology, Guangzhou, China\\

    \textsuperscript{\rm 3}School of Science and Engineering, The Chinese University of Hong Kong (Shenzhen), Shenzhen, China\\

    \{yudonghu, jinkeren\}@cuhk.edu.cn,
    ftyuejuhan@mail.scut.edu.cn,
    ruisun@link.cuhk.edu.cn
}
\nocopyright

\usepackage{bibentry}

\begin{document}

\maketitle

\begin{abstract}
Capsule Network (CapsNet) has demonstrated significant potential in visual recognition by capturing spatial relationships and part-whole hierarchies for learning equivariant feature representations. However, existing CapsNet and variants often rely on a single high-level feature map, overlooking the rich complementary information provided by multi-scale features. Furthermore, conventional feature fusion strategies, such as addition and concatenation, struggle to reconcile multi-scale feature discrepancies, leading to suboptimal classification performance. To address these limitations, we propose the Multi-Scale Patchify Capsule Network (MSPCaps), a novel architecture that integrates multi-scale feature learning and efficient capsule routing. Specifically, MSPCaps consists of three key components: a Multi-Scale ResNet Backbone (MSRB), a Patchify Capsule Layer (PatchifyCaps), and a  Cross-Agreement Routing (CAR) block. First, the MSRB extracts diverse multi-scale feature representations from input images, preserving both fine-grained details and global contextual information. Second, the PatchifyCaps partitions these multi-scale features into primary capsules using a uniform patch size, equipping the model with the ability to learn from diverse receptive fields. Finally, the CAR block adaptively routes the multi-scale capsules by identifying cross-scale prediction pairs with maximum agreement. Unlike the simple concatenation of multiple self-routing blocks, CAR ensures that only the most coherent capsules (best part-to-whole pairs) contribute to the final voting. Our proposed MSPCaps achieves remarkable scalability and superior robustness, consistently surpassing multiple baseline methods in terms of classification accuracy, with configurations ranging from a highly efficient Tiny model (344.3K parameters) to a powerful Large model (10.9M parameters), highlighting its potential in advancing feature representation learning. The code is available at \textit{https://github.com/abdn-hyd/MSPCaps}.
\end{abstract}


\section{Introduction}
Convolutional Neural Networks (CNNs) have proven exceptional capabilities in extracting hierarchical image representations. However, understanding visual scenes requires not only detecting parts or objects, but also understanding their spatial relationships and part-whole structures. The inability of CNNs to inherently model these relationships leads to significant limitations, including low robustness and poor spatial object correlation. To address these issues, Capsule Network (CapsNet), first introduced by \cite{DR2017}, encodes entities as vectors representing intrinsic properties, such as pose and orientation. In simple terms, CNNs determine if a feature is present, while CapsNet explains how it exists by capturing the relationships between parts.

\begin{figure}[t]
\centering
\includegraphics[width=\columnwidth]{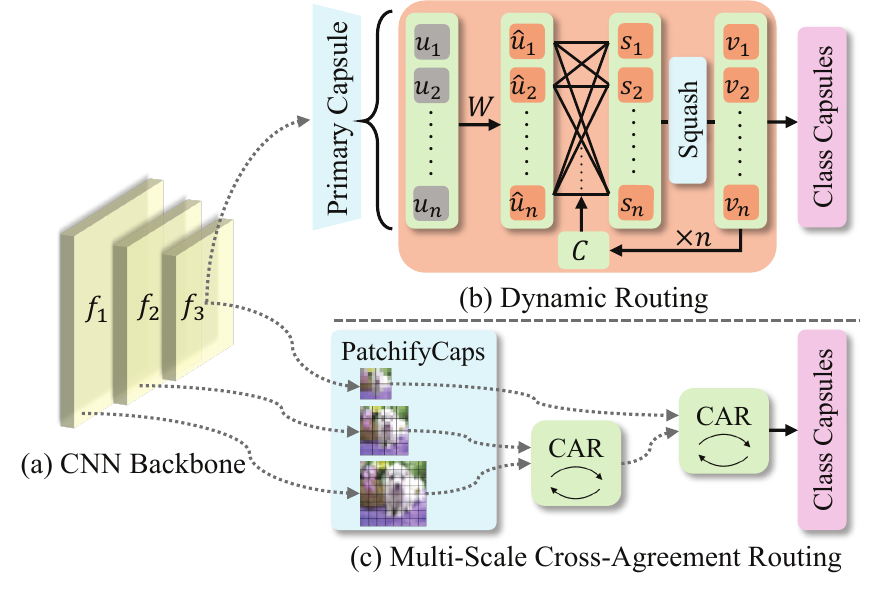}
\caption{Comparative overview of conventional CapsNet and our proposed MSPCaps. The path (a) $\rightarrow$ (b) shows the conventional CapsNet with dynamic routing, where the prediction capsules $\hat{u}$ are first generated by applying weight matrices $W$ on primary capsules $u$, followed by an iterative routing-by-agreement process to update the coupling coefficients $C$ for $n$ iterations. The class capsules are then produced in the last iteration based on output capsules $v$. In contrast, the path (a) $\rightarrow$ (c) shows the proposed MSPCaps, which leverages multi-scale features to generate initial capsules and employs a non-iterative cross-agreement routing mechanism for voting to compute output capsules.}
\label{Contrast}
\end{figure}

At its core, CapsNet operates on the principle of hierarchical composition. At the very bottom, primary capsules identify basic visual elements called parts and encode each part's attributes. Then, they actively predict the presence and properties of high-level entities they may form using learned transformation matrices. In practice, this predictive voting procedure is orchestrated by an iterative dynamic routing mechanism. Low-level capsules exchange votes and update their coupling coefficients based on the agreement achieved with the predicted capsules. As a result, only those high-level capsules receiving mutual agreement from multiple sources are strongly activated.

Recent studies have advanced capsules with deep backbones \cite{DeepCaps, DA-CapsNet, orthcaps} or attention routing \cite{Mazzia2021, AttnRouting2019, AA-CapsNet}. However, despite their compelling designs, CapsNet and its variants still face several limitations: (1) Existing studies \cite{Mazzia2021} instantiate primary capsules directly from entire global feature maps. This runs counter to the intuition that the lowest-level capsule should focus on simple visual primitives that are spatially localized; (2) Existing capsule models mainly rely on single-resolution features, neglecting the rich complementary information available across different scales; (3) Conventional feature fusion strategies, such as simple addition and concatenation, struggle to effectively reconcile the discrepancies between multi-scale features.

To address these limitations, we propose the Multi-Scale Patchify Capsule Network (MSPCaps), a novel architecture that integrates multi-scale feature learning and efficient capsule routing.  Fig.~\ref{Contrast} presents a comparative overview of the conventional CapsNet and our MSPCaps model. Specifically, MSPCaps consists of three core components: a Multi-Scale ResNet Backbone (MSRB), a Patchify Capsule Layer (PatchifyCaps), and a Cross-Agreement Routing (CAR) block. Firstly, MSRB utilizes a shared ResNet backbone to extract a hierarchy of multi-scale features from input images. Typically, images with different resolutions contain complementary information: high-resolution features tend to preserve fine-grained details (local primitives), while low-resolution representations provide global semantic information (object-level entities). Based on this, PatchifyCaps takes the feature maps from different scales and partitions each feature map into a grid of fixed-sized, non-overlapping patches, generating spatially localized primary capsules. This mechanism not only enables the model to focus on local regions but also allows it to learn from diverse receptive fields, which capture intricate patterns at high resolutions and broader abstractions at lower resolutions. Finally, to effectively fuse multi-scale features, the CAR block aggregates multi-scale capsules by identifying cross-scale prediction pairs with maximum agreement, thereby ensuring that only the most coherent capsules (best part-to-whole pairs) contribute to the final decision. Moreover, by progressively fusing high and low resolution capsules, CAR provides rich guidance for generating complex higher-level capsules and acts as an implicit regularizer, preventing the model from being misled by excessive fine-grained details. The effectiveness of the proposed MSPCaps is evaluated on four baseline datasets: MNIST, FashionMNIST, CIFAR-10, and SVHN. The results show that MSPCaps achieve state-of-the-art performance and superior robustness compared to multiple baselines within capsule networks.

Our main contributions are summarized as follows:
\begin{itemize} 
    \item We introduce a novel MSRB and a PatchifyCaps layer in MSPCaps to extract and generate multi-scale capsules, enabling learning from diverse receptive fields while preserving fine-grained details and global context. 
    \item  We propose a novel CAR, an adaptive routing mechanism that effectively fuses multi-scale capsules by selecting pairs with maximum agreement, thereby ensuring that only coherent capsules contribute to decisions.
    \item Experimental results demonstrate that our method consistently outperforms multiple baseline methods in classification accuracy, with configurations ranging from an efficient tiny model (344.3K parameters) to a powerful large model (10.9M parameters). Moreover, the large model also shows significant robustness against adversarial attacks and notably surpasses CapsNet.
\end{itemize} 

\section{Related Works}
\begin{figure*}[t]
\centering
\includegraphics[width=1.0\textwidth]{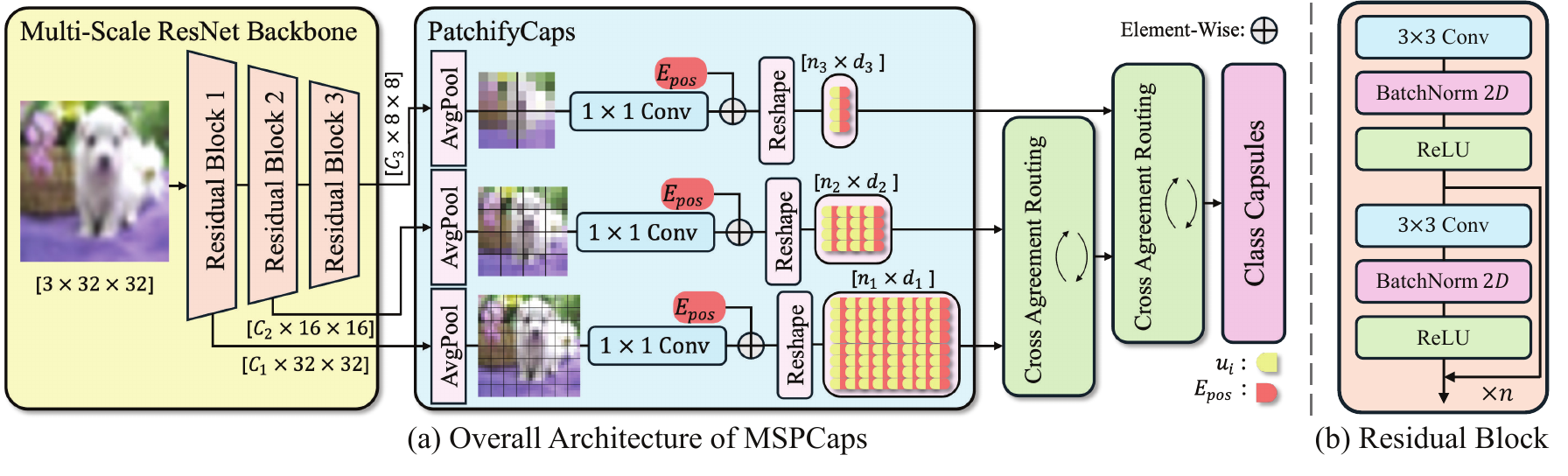}
\caption{(a) The overall architecture of MSPCaps, which consists of an MSRB, a PatchifyCaps, and two CAR blocks. The backbone is designed with three residual blocks to generate features at varying scales. (b) Each residual block starts with a downsampling convolution (stride=2), followed by $n$ identical convolutional modules, each with a residual connection.}
\label{MainModel}
\end{figure*}

\textbf{Routing Techniques in Capsule Networks.} Research on CapNets has explored various routing strategies to improve efficiency and representation capabilities. Early methods such as Dynamic Routing (DR) \cite{DR2017} and EM Routing \cite{EMRouting} employ iterative clustering mechanisms to refine coupling coefficients, using vector or matrix representations, respectively. 
More recently, attention-based \cite{AttnRouting2019, DA-CapsNet} and non-iterative \cite{GR, parsecaps} strategies have gained prominence. Specifically, Efficient-Caps \cite{Mazzia2021} introduces a self-attention routing scheme to compute coupling coefficients based on similarity among capsule votes, enabling fully parallel and non-iterative routing. OrthCaps \cite{orthcaps} further enhances efficiency by integrating pruning and orthogonal sparse attention routing, effectively reducing redundancy while maintaining accuracy. Moreover, SR-CapsNet \cite{SR} draws inspiration from the Mixture-of-Experts (MOE) architecture, using two sub-networks per capsule to generate pose vectors and activations, achieving expert-style routing. However, those methods predominantly rely on single-resolution features, and little attention has been paid to multi-scale routing strategies.

\textbf{Multi-Scale Feature and Fusion.} Multi-scale feature modeling has been widely explored to improve visual classification by capturing both local details and global context. CNN-based methods \cite{FPN, HRNet, res2net, Multi-dense} incorporate multi-scale information and fuse them following a common pattern during fusion: combining the features via concatenation or addition, followed by projection through convolutions. Similarly, CapsNet variants such as MS-CapsNet \cite{MSCaps} and RS-CapsNet \cite{RS-Caps} fuse capsules across scales by concatenating and jointly routing multi-scale capsules, following a comparable fusion paradigm. Nevertheless, the multi-scale fusion in CapNets remains shallow, as most existing approaches concatenate or add features across scales without deeper interaction, limiting the effectiveness of feature integration.


\section{Methodology}
We propose MSPCaps, a novel architecture constructed from three core modules: (1) a Multi-Scale ResNet Backbone (MSRB), designed to capture both fine-grained details and global semantic information; (2) a Patchify Capsule Layer (PatchifyCaps), which learns complementary information across different receptive fields and transforms the extracted features into primary capsules; (3) and a Cross-Agreement Routing (CAR) block for adaptive cross-scale capsule voting. To demonstrate the scalability of the proposed architecture, we present two model versions: a lightweight model--MSPCaps-T and a powerful model--MSPCaps-L. Both models share the same core design and differ only in key hyperparameters and the utilization of the weight-sharing mechanism within CAR.

\subsection{Overall Architecture}
As shown in Fig.~\ref{MainModel}, the MSPCaps architecture first processes an input image $x$ using MSRB, extracting hierarchical feature maps at three distinct scales, $f_i \in \mathbb{R}^{C_i \times H_i \times W_i}$ for $i \in \{1, 2, 3\}$. Each subsequent scale halves the spatial dimensions, i.e., $H_{i+1} = H_i/2$. These multi-scale features are then fed into PatchifyCaps, which partitions each feature map $f_i$ into non-overlapping patches of size $p \times p$ and applies a $1 \times 1$ convolution to generate an initial set of primary capsules $u_i \in \mathbb{R}^{n_i \times d_i}$. Here, $n_i = (H_i \cdot W_i)/p^2$ denotes number of capsules, and $d_i$ is the capsule dimension.

To integrate information across scales for voting, we propose a hierarchical fusion strategy powered by the CAR mechanism. Instead of using a simple concatenation of multiple self-routing modules, our approach fuses capsule sets sequentially in a fine-to-coarse manner. Specifically, the process begins with the capsules from the finest resolution level, $u_1$. The first CAR block uses these capsules to query the mid-level capsules $u_2$, producing a set of intermediate fused capsules $u_{1-2}$. Then, a second CAR block uses this fused representation to query the capsules from the coarsest resolution level $u_{3}$. This progressive fusion process effectively distills rich, multi-scale contextual information while preserving complementary details at each resolution level. The final routing step outputs a set of class capsules, which encapsulate comprehensive multi-scale information for downstream prediction tasks.

\subsection{Multi-Scale ResNet Backbone}
The MSRB is designed to generate hierarchical features. Inspired by the progressive downsampling strategy in ResNet \cite{ResNet}, MSRB first generates the finest feature map $u_1$ and progressively reduces spatial resolution using convolutions with a kernel size of 3 and a stride of 2. Deeper feature learning is achieved using subsequent convolutions with a kernel size of 3 and a stride of 1.

The primary motivation for this design is to control the learning process at different levels of abstraction and align it with the subsequent PatchifyCaps. For high-resolution (fine-grained) feature maps, we employ a shallower network with fewer parameters, encouraging the model to capture essential local details while reducing the risk of overfitting caused by overly complex fine-grained features. Conversely, for low-resolution (coarse-grained) feature maps, which encapsulate global context, we utilize a deeper network with richer parameterization to enhance abstract representations learning. Furthermore, this structure introduces a potent regularization effect that stems from the use of a hierarchical weight-sharing mechanism. Specifically, the initial residual blocks are shared across all scales, while deeper blocks are specialized for coarse resolutions. This dual supervision updates the shared kernels with gradients from both the fine and coarse outputs, which implicitly regularizes the network to extract more general and robust representations.

\begin{figure*}[t]
\centering
\includegraphics[width=0.95\textwidth]{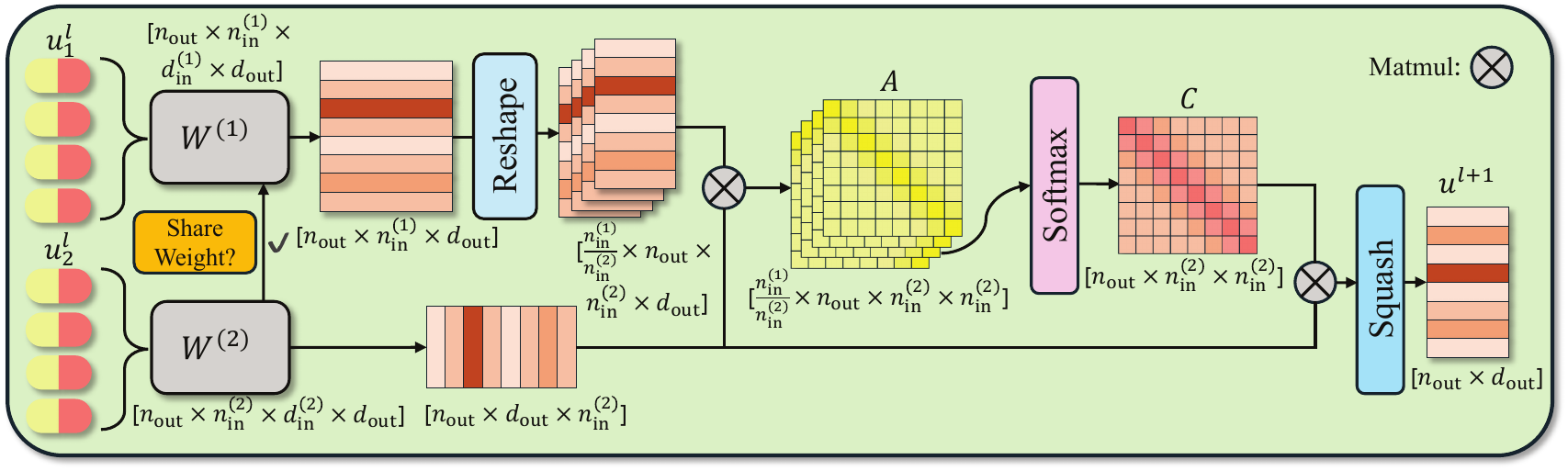}
\caption{An illustration of the cross-agreement routing block, where $A$ represents the raw agreement scores matrix and $C$ denotes the final coupling coefficients matrix.}
\label{CAR}
\end{figure*}

\subsection{Patchify Capsule Layer}
A key challenge in multi-scale feature representation lies in preserving both critical fine-grained and global information. To address this, we propose PatchifyCaps, a module that interfaces with MSRB to generate multi-scale capsules. Inspired by patch-based architectures such as Vision Transformers \cite{vit}, PatchifyCaps partitions multi-scale feature maps into a grid of patches. By applying a uniform patch size to feature maps of different scales, the capsules derived from high-resolution maps capture local patterns via small receptive fields, while those from coarser maps gain a wider view to model global structures. Beyond creating primary capsules, PatchifyCaps intrinsically links receptive field size to the feature hierarchy, enabling adaptive perception across scales.

To implement this, PatchifyCaps employs a ``one-patch, one-capsule" mapping. It first partitions each feature map into a grid of non-overlapping patches using an average pooling layer. This partitioning strategy enforces a strict local receptive field, ensuring each resulting capsule is sensitive to spatial patterns within its designated region. Subsequently,  a $1 \times 1$ convolution adapts the feature channels to the desired capsule dimension, $d_i$. The resulting tensor is then reshaped into a sequence of capsules, $u_i \in \mathbb{R}^{n_i \times d_i}$. To compensate for the spatial context lost during the flattening step, we augment each capsule vector with a learnable positional embedding. This step ensures that each capsule retains its region-sensitive properties, encoding both the presence and location of features on the feature map grid. The detailed calculation process is summarized in Alg.~\ref{alg:patchify}.

\begin{algorithm}[tb]
\caption{PatchifyCaps}
\label{alg:patchify}
\textbf{Input}: Feature map $f_i \in \mathbb{R}^{C_i \times H_i \times W_i}$, patch size $p$. \\
\textbf{Parameter}: $W_{\text{conv}}$, $E_{\text{pos}} \in \mathbb{R}^{n_i \times d_i}$.\\
\textbf{Output}: $u_i \in \mathbb{R}^{n_i \times d_i}$.
\begin{algorithmic}[1] 
    \STATE $u_{\text{patchify}} \leftarrow \text{Avgpool2d}(f_i, k=p, s=p)$.
    \STATE $u_{\text{unshaped}} \leftarrow \text{Conv2d}(u_{\text{patchify}}, W_{\text{conv}}, k=1, s=1, in=C_i, out=d_i)$.
    \STATE Let $n_i = \frac{H_i}{p} \times \frac{W_i}{p}$.
    \STATE $u_{\text{flat}} \leftarrow \text{Reshape}(u_{\text{unshaped}}, (n_i, d_i))$.
    \STATE $u_i \leftarrow u_{\text{flat}} + E_{\text{pos}}$.
    \STATE $u_i \leftarrow \text{LayerNorm}(u_i)$.
    \STATE \textbf{return} $u_i$
\end{algorithmic}
\end{algorithm}

The ``one-patch, one-capsule" approach is computationally efficient. By replacing parallel convolutions with a simple pooling layer and a single $1 \times 1$ convolution, it significantly reduces the number of parameters required to form the primary capsules. Moreover, we can effectively compress the number of primary capsules through an appropriate patch size. For instance, when $p=4$, the number of primary capsules is reduced to just 84 compared to 1152 in \cite{DR2017}, with a decrease of over 90\%. This drastic reduction proportionally decreases the parameters and computations needed for weight projection during voting, yielding a substantial downstream effect.

\subsection{Cross-Agreement Routing}
The efficacy of conventional capsule routing is often limited by its single-source design, which fails to leverage the rich, multi-scale information essential for complex feature voting. Although prior works like RS-CapsNet \cite{RS-Caps} incorporate capsules from various scales, their fusion remains simplistic, merely aggregating multiple intermediate capsules for final routing. To overcome this limitation, we introduce CAR, a novel routing mechanism to achieve sophisticated agreement-based voting across feature scales. As shown in Fig.~\ref{CAR}, the CAR begins with two parallel input capsules from a lower layer $l$: finer capsules $u_1^l \in \mathbb{R}^{n_{\text{in}}^{(1)} \times d_{\text{in}}^{(1)}}$ representing smaller objects, and coarser capsules $u_2^l \in \mathbb{R}^{n_{\text{in}}^{(2)} \times d_{\text{in}}^{(2)}}$ containing larger ones. Both capsules are first transformed into prediction capsules $\hat{u}^{(1)}$ and $\hat{u}^{(2)}$ for the $n_{\text{out}}$ output capsules in layer $l+1$. This process is achieved by applying transformation matrices, $W^{(1)} \in \mathbb{R}^{n_{\text{out}} \times n_{\text{in}}^{(1)} \times d_{\text{in}}^{(1)} \times d_{\text{out}}}$ and $W^{(2)} \in \mathbb{R}^{n_{\text{out}} \times n_{\text{in}}^{(2)} \times d_{\text{in}}^{(2)} \times d_{\text{out}}}$, which can be given by
\begin{equation}
    \hat{u}_{j|i}^{(1)} = W_{j,i}^{(1)} u_{1,i}^{l} \quad \text{and} \quad \hat{u}_{j|k}^{(2)} = W_{j,k}^{(2)} u_{2,k}^{l}
    \label{eq:proj}
\end{equation}
where $\hat{u}_{j|i}^{(1)}$ and $\hat{u}_{j|k}^{(2)}$ denote the vote from capsule $i$ and $k$ for output capsule $j$.
\begin{algorithm}[tb]
\caption{Cross-Agreement Routing (CAR)}
\label{alg:CAR}
\textbf{Input}: $u_1^l \in \mathbb{R}^{n_{\text{in}}^{(1)} \times d_{\text{in}}^{(1)}}$, $u_2^l \in \mathbb{R}^{n_{\text{in}}^{(2)} \times d_{\text{in}}^{(2)}}$. \\
\textbf{Parameter}: $W^{(1)} \in \mathbb{R}^{n_{\text{out}} \times n_{\text{in}}^{(1)} \times d_{\text{in}}^{(1)} \times d_{\text{out}}}$, $W^{(2)} \in \mathbb{R}^{n_{\text{out}} \times n_{\text{in}}^{(2)} \times d_{\text{in}}^{(2)} \times d_{\text{out}}}$.\\
\textbf{Output}: $u^{l+1} \in \mathbb{R}^{n_{\text{out}} \times d_{\text{out}}}$.
\begin{algorithmic}[1] 
    \STATE $\hat{\mathbf{U}}^{(2)} \leftarrow W^{(2)} u_2^l$. \hfill $\triangleright$ Voting
    \IF{shared weight}
        \STATE $\hat{\mathbf{U}}^{(1)} \leftarrow W^{(2)} u_1^l$. \hfill $\triangleright$ Voting
    \ELSE
        \STATE $\hat{\mathbf{U}}^{(1)} \leftarrow W^{(1)} u_1^l$. \hfill $\triangleright$ Voting
    \ENDIF
    \STATE $\mathbf{A} \leftarrow \max_{\text{groups}} \left( \frac{\hat{\mathbf{U}}^{(1)} (\hat{\mathbf{U}}^{(2)})^T}{\sqrt{d_{\text{out}}}} \right)$. \hfill $\triangleright$ Agreement Scores
    \STATE $\mathbf{C} \leftarrow \text{Softmax}(\mathbf{A})$. \hfill $\triangleright$ Coupling Coefficients
    \STATE $\mathbf{V} \leftarrow \mathbf{C}  \hat{\mathbf{U}}^{(2)}$.
    \STATE $v_j \leftarrow \sum_{k} \mathbf{V}_{j,k}$, where $j=1, \cdots, n_{\text{out}}$.
    \STATE $u_j^{l+1} \leftarrow \text{squash}(v_j)$.
    \STATE \textbf{return} $u_j^{l+1}$
\end{algorithmic}
\end{algorithm}

The foundation of our routing mechanism is based on a key hypothesis: The uniform patch size for cross-scale capsules ensures inherent spatial correspondence. Specifically, a local group of $s = n_{\text{in}}^{(1)} / n_{\text{in}}^{(2)}$ capsules from the finer input $u_1^{l}$ corresponds spatially to a single capsule from the coarser input $u_2^{l}$, resembling the parts-to-whole relationships in traditional dynamic routing. Therefore, capsules within the same local region should exhibit high similarity and thus make a consistent contribution for voting. Building on this premise, we also introduce a lightweight design (weight-sharing mechanism) to reduce parameters  for MSPCaps-T by assuming that the mutual information between two scales of capsules is sufficient for classification. To implement this, we simply apply transformation matrix $W^{(2)}$ to the finer capsules $u_1^l$ to generate prediction capsules $\hat{u}^{(1)}$, which is given by
\begin{equation}
    \hat{u}_{j|m,k}^{(1)} = W_{j,k}^{(2)} u^{l}_{1,m,k}
    \label{eq:proj_shared}
\end{equation}
where $u^{l}_{1,m,k}$ denotes the $m$-th capsule within the $k$-th spatial group of $u_1^{l}$. This significantly reduces the number of parameters by eliminating the use of $W^{(1)}$. 

To perform cross-agreement routing, we first reshape the prediction vectors $\hat{u}^{(1)}$ and organize them into $n_{\text{in}}^{(2)}$ groups, each containing $s$ vectors. The agreement score between the $k$-th capsule group and the $j$-th output capsule is determined by the maximum similarity found within that group, which is computed as
\begin{equation}
    c_{jk} = \text{Softmax} \left( \max_{m=1,\dots,s} \left( \frac{(\hat{u}_{j|m, k}^{(1)}) \cdot (\hat{u}_{j|k}^{(2)})^T}{\sqrt{d_{\text{out}}}} \right) \right)
    \label{eq:agreement}
\end{equation}
where $\{\hat{u}_{j|m,k}^{(1)}\}_{m=1}^s$ is the group of prediction vectors from the finer capsules that spatially correspond to the $k$-th coarser capsule. By adaptively selecting the maximum similarity, the most representative feature alignment (best part-to-whole pair) within the local patch guides the routing process. The raw agreement scores are then normalized using the softmax function to compute the final coupling coefficients, $c_{jk}$. The resulting output capsule $v_j$ is then calculated as a weighted sum over the coarser-scale prediction vectors, $v_j = \sum_k c_{jk} \hat{u}_{j|k}^{(2)}$, which is then passed through a squash activation function \cite{DR2017} to generate the intermediate capsules or class capsules.

\section{Experiments}
\subsection{Datasets and Evaluation Metrics}
In our experiments, we select four baseline datasets: MNIST \cite{lecun1998mnist}, FashionMNIST \cite{xiao2017fashion}, SVHN \cite{SVHN}, and CIFAR-10 \cite{Cifar}, which are widely used in the literature of capsule networks. During data preprocessing, the data samples in MNIST and Fashion-MNIST are resized from $28 \times 28$ to $32 \times 32$ for a consistent comparison using the same model configuration. To enhance generalization, we apply data augmentation to the training splits only. Specifically, the images in Fashion-MNIST, SVHN, and CIFAR-10 undergo a $32 \times 32$ random crop with 4-pixel padding on each side, followed by a random horizontal flip except for SVHN. For the MNIST and SVHN datasets, we apply random rotations within $\pm 15$ degrees. 

To comprehensively assess the proposed model, we employ the classification accuracy as the evaluation metric, which is defined as the proportion of samples where the class capsule with the largest L2 norm corresponds to the true class label. Additionally, we take the total number of trainable parameters (Params) as a crucial metric, indicating the computational complexity of each model.

\subsection{Implementation Details}
All experiments are conducted using PyTorch version 2.5.1 with Python 3.12 and CUDA 12.4 on a single NVIDIA RTX 4090D GPU. In particular, we utilize the margin loss as defined in \cite{DR2017}. The reconstruction loss is omitted as it provides only a negligible improvement in model performance. We train our model for 300 epochs using a batch size of $128$. Additionally, we employ the AdamW optimizer with a learning rate of $5 \times 10^{-4}$ and a weight decay of $1 \times 10^{-4}$. We also use a learning rate schedule with a 5-epoch linear warmup from 10\% of the base value, followed by a cosine annealing decay for the remaining $295$ epochs to a minimum of $1 \times 10^{-6}$. Other hyperparameters of our model variants are provided in Tab.~\ref{tab:hyperparameter}.
\begin{table}[t]
\centering
\small
\setlength{\tabcolsep}{1.8mm}
\begin{tabular}{c c c}
\hline\hline
\textbf{Hyperparameter} & \textbf{MSPCaps-T} & \textbf{MSPCaps-L} \\
\hline 
$C_1,C_2,C_3$ & 32, 64, 128 & 128, 256, 512 \\
$N$ & 2 & 3 \\
$d_1,d_2,d_3$ & 8, 8, 16 & 16, 32, 64 \\
$d_{\text{mid}}$ & 16 & 64 \\
$d_{\text{out}}$ & 32 & 128 \\
$p$ & 4 & 4 \\
Weight Shared? & \checkmark & \ding{55} \\
\hline\hline %
\end{tabular}
\caption{Hyperparameters of our model variants. $N$: number of convolutions in each residual block; $d_{\text{mid}}$: intermediate capsules dimension; $d_{\text{out}}$: class capsules dimension.}
\label{tab:hyperparameter}
\end{table}

\subsection{Baselines}
To demonstrate the advantage of the proposed MSPCaps, we compare it with 9 state-of-the-art models, including: (1) CapsNet \cite{DR2017}; (2) AA-CapsNet \cite{AA-CapsNet}; (3) AR-CapsNet \cite{AttnRouting2019}; (4) DA-CapsNet \cite{DA-CapsNet}; (5) DeepCaps \cite{DeepCaps}; (6) RS-CapsNet \cite{RS-Caps}; (7) CapsNet with inverted dot-product (IDP-CapsNet) \cite{DotCaps}; (8) Efficient-Caps \cite{Mazzia2021}; and (9) OrthCaps \cite{orthcaps}.

\subsection{Comparison Results}
Tab.~\ref{tab:MSPCaps-T Performance} and Tab.~\ref{tab:MSPCaps-L Performance} illustrate the 
performance of MSPCaps-T and MSPCaps-L against the baseline models. The results for each model are based on the best performance achieved during training with random seed initialization. The asterisk (*) indicates that the result is reported directly from the original paper, as official code is unavailable.

From Tab.~\ref{tab:MSPCaps-T Performance}, we can see that our MSPCaps-T demonstrates exceptional performance while maintaining a relatively low parameter count, although it is not explicitly designed for parameter reduction. Specifically, it achieves accuracies of 99.69\% on MNIST, 95.79\% on SVHN, and 88.71\% on CIFAR-10, consistently outperforming multiple baseline models except OrthCaps-S on SVHN. An insightful comparison is made with OrthCaps-S and Efficient-Caps, which both have a smaller number of parameters. While MSPCaps-T utilizes approximately 240K more parameters than OrthCaps-S, this increase yields a notable performance trade-off across datasets. On SVHN, OrthCaps-S holds an advantage with a gap of 0.47\%. However, the true strength of our model's architecture becomes evident on the more challenging CIFAR-10 dataset. Here, MSPCaps-T achieves a new state-of-the-art accuracy of 88.71\%, surpassing OrthCaps-S by a significant margin of 1.87\%. Additionally, compared with larger models such as DA-CapsNet and AA-CapsNet, our model demonstrates significant improvement while being over 20 times more parameter-efficient.
\begin{table}[t]
\centering
\small
\setlength{\tabcolsep}{2.0mm}
\begin{tabular}{l c c c c}
\hline\hline
\textbf{Networks} & \textbf{Params} & \textbf{MNIST} & \textbf{SVHN} & \textbf{CIFAR-10} \\
\hline
CapsNet                          & 6.8M & 99.45 & 93.96 & 74.88 \\
DA-CapsNet                       & 7M* & 99.53* & 94.82* & 85.47* \\
AA-CapsNet                       & 6.6M* & 99.34* & 92.13* & 71.60* \\
IDP-CapsNet                      & 560K & 99.35 & - & 85.17 \\
Efficient-Caps                   & 162.4K & 99.65 & 95.20 & 77.81 \\
OrthCaps-S                       & \textbf{105.5K*} & 99.68* & \textbf{96.26*} & 86.84* \\
\hline
MSPCaps-T                        & 344.3K & \textbf{99.69} & 95.79 & \textbf{88.71} \\
\hline\hline %
\end{tabular}
\caption{Performance comparison of MSPCaps-T with baseline models on MNIST, SVHN and CIFAR-10.}
\label{tab:MSPCaps-T Performance}
\end{table}

\begin{table}[t]
\centering
\small
\setlength{\tabcolsep}{1mm}
\begin{tabular}{l c c c c}
\hline\hline
\textbf{Networks} & \textbf{Params} & \textbf{MNIST} & \textbf{FashionMNIST} & \textbf{CIFAR-10} \\
\hline
CapsNet$^{\dag}$           & 6.8M* & - & - & 89.40* \\
RS-CapsNet                 & 5.0M* & - & 93.51* & 89.81* \\
AR-CapsNet                       & 9.6M* & 99.46* & - & 88.61* \\
AR-CapsNet$^{\dag}$        & 9.6M* & 99.49* & - & 90.11* \\
DeepCaps                   & 13.5M* & 99.72* & 94.46* & 91.01* \\
OrthCaps-D                 & \textbf{574K*} & 99.58* & 94.60* & 90.56* \\
\hline
MSPCaps-L                  & 10.9M & \textbf{99.73} & \textbf{95.05} & \textbf{92.88} \\
\hline\hline %
\end{tabular}
\caption{Performance comparison of MSPCaps-L with baseline models on MNIST, FashionMNIST and CIFAR-10. $^{\dag}$ indicates the result is obtained using an ensemble of 7 models.}
\label{tab:MSPCaps-L Performance}
\end{table}

\begin{table}[t]
\centering
\small
\setlength{\tabcolsep}{1.5mm}
\begin{tabular}{c c c c c}
\hline\hline
\textbf{32 $\times$ 32?} & \textbf{16 $\times$ 16?}& \textbf{8 $\times$ 8?} & \textbf{Params} & \textbf{CIFAR-10} \\
\hline
\checkmark &  \ding{55} & \ding{55} & 92.9K & 74.81 \\
\ding{55} & \checkmark & \ding{55} & 86.9K & 81.90 \\
\ding{55} & \ding{55} & \checkmark & 293.6K & 87.48 \\
\checkmark &  \checkmark & \ding{55} & 87.6K &  81.57 \\
\hline
\checkmark & \checkmark & \checkmark & 344.3K & \textbf{88.71} \\
\hline\hline
\end{tabular}
\caption{Ablation results of MSPCaps-T model on CIFAR-10, evaluating the impact of different feature map scales.}
\label{tab:Scale_Ablation}
\end{table}

To fully explore the potential of our architecture under parameter-sufficient settings, we compare the full-sized model--MSPCaps-L with other deeper baselines. As shown in Tab.~\ref{tab:MSPCaps-L Performance}, MSPCaps-L exhibits the best performance on all datasets, specifically achieving accuracies of 99.73\% on MNIST, 95.05\% on FashionMNIST, and a notable 92.88\% on CIFAR-10. In the category of deep CapsNet, our MSPCaps-L demonstrates a significant performance gain compared to DeepCaps by 1.87\% on CIFAR-10. This can be attributed to the powerful multi-scale architecture. The improvement is further enhanced when compared to the smaller models, i.e., CapsNet$^\dag$ and AR-CapsNet$^{\dag}$. The increase in parameter counts for our MSPCaps-L results in a significant 3.48\% and 2.77\% accuracy improvement on CIFAR-10, respectively. Furthermore, compared to the multi-scale based RS-CapsNet, our model outperforms with a significant gap of 3.07\% on CIFAR-10 which demonstrates the superiority of our architecture. These results validate our model's scalability and the ability to convert an increased parameter budget into robust feature representation.

\subsection{Ablation Study}
\textbf{Contribution Analysis of Multi-Scale Features}. To demonstrate the contribution of multi-scale capsules for final prediction, we compare our full MSPCaps-T model against several tiny ablated variants on CIFAR-10, as shown in Tab~\ref{tab:Scale_Ablation}. These variants use capsules generated from specific feature map scales ($32 \times 32$, $16 \times 16$, and $8 \times 8$). In the single-scale experiments, we input two identical sets of capsules, adopting the same capsule dimension configuration as the first CAR block in MSPCaps-T. The results affirm the effectiveness of our multi-scale design, with the full model achieving a top accuracy of 88.71\%. The $8\times8$ feature map acts as the primary contributor, establishing the most robust semantic foundation with its global contextual view compared to other ablated variants. However, this global view lacks information on fine details, which is addressed by the high-resolution $32 \times 32$ map. By supplying essential fine-grained details for voting, it helps the model resolve ambiguities in classification. Although this detail-oriented contribution is insufficient by itself, it is still indispensable for refining the model's predictions. 
\begin{table}[t]
\centering
\small
\setlength{\tabcolsep}{3.3mm}
\begin{tabular}{l c c c}
\hline\hline
\textbf{Variants} & \textbf{Routing} & \textbf{Params} & \textbf{CIFAR-10} \\
\hline
\multirow{2}{*}{MSPCaps-T} & ($4 \times $DR) & 483.5K & 87.46 \\
                           & ($2 \times $CAR) & 344.3K & \textbf{88.71} \\
\midrule
\multirow{2}{*}{MSPCaps-L} & ($4 \times $DR) & 14.6M & 92.56 \\
                           & ($2 \times $CAR) & 10.9M & \textbf{92.88} \\
\hline\hline
\end{tabular}
\caption{Ablation study of different routing algorithms for MSPCaps. DR denotes dynamic routing.}
\label{tab:Routing_Ablation}
\end{table}

\begin{table}[t]
\centering
\small
\setlength{\tabcolsep}{2.1mm}
\begin{tabular}{l c c c}
\hline\hline
\textbf{Variants} & \textbf{Shared Weights?} & \textbf{Params} & \textbf{CIFAR-10} \\
\hline
\multirow{2}{*}{MSPCaps-T} & \checkmark & 344.3K & \textbf{88.71} \\
                           & \ding{55} & 557.3K & 87.59 \\
\midrule
\multirow{2}{*}{MSPCaps-L} & \checkmark & 7.9M & 92.50 \\
                           & \ding{55} & 10.9M & \textbf{92.88} \\
\hline\hline
\end{tabular}
\caption{Ablation study of weight sharing 
within CAR.}
\label{tab:Shared_Ablation}
\end{table}

\begin{table}[t]
\centering
\small
\setlength{\tabcolsep}{1.4mm}
\begin{tabular}{l c c c c}
\hline\hline
\textbf{Variants} & \textbf{Patch Size} & \textbf{Num Caps} & \textbf{Params} & \textbf{CIFAR-10} \\
\hline
\multirow{3}{*}{MSPCaps-T} & 2 & 336 & 899.3K & 88.01 \\
                           & 3 & 129 & 388.7K & 86.34 \\
                           & 4 & 84 & 344.3K & \textbf{88.71} \\
\midrule
\multirow{3}{*}{MSPCaps-L} & 2 & 336 & 39.4M & 93.16 \\
                           & 3 & 129 & 13.7M & \textbf{93.30} \\
                           & 4 & 84 & 10.9M & 92.88 \\
\hline\hline
\end{tabular}
\caption{Analysis of different patch sizes for PatchifyCaps.} 
\label{tab:PatchSize}
\end{table}

\textbf{Quantitative Comparison between DR and CAR}. 
To validate the proposed CAR block, we compare it with the original dynamic routing (DR) \cite{DR2017} on both MSPCaps-T and MSPCaps-L. The ablated variants replace the CAR blocks with DR blocks, where three routing blocks first process the multi-scale features independently, after which their outputs are concatenated and passed to a final, fourth routing block. As shown in Tab.~\ref{tab:Routing_Ablation}, the proposed CAR block demonstrates superior performance and scalability across both model variants. For MSPCaps-T, our CAR variant significantly surpasses DR while reducing parameters by 28.7\%. This advantage is further amplified on MSPCaps-L, where our CAR variant achieves the highest accuracy with a 25\% parameter reduction compared to DR. These results highlight CAR's effectiveness in fusing cross-scale capsules for voting, offering consistent gains in both accuracy and parameter efficiency.

\textbf{Share Weight Mechanism Analysis}. 
Finally, we conduct an ablation study on the effect of weight sharing within the CAR block. The results are shown in Tab.~\ref{tab:Shared_Ablation}. We can observe an interesting trade-off depending on model capacity. Specifically, for the lightweight MSPCaps-T, weight sharing proves to be an effective regularization strategy, enhancing mutual information utilization in parameter-constrained scenarios.  In contrast, for the larger MSPCaps-L, disabling weight sharing improves the ``part-to-whole" voting process by leveraging its ample parameters.

\subsection{Patch Size Analysis}
To determine the optimal patch size $p$, we evaluate classification accuracy on the CIFAR-10 dataset with patch sizes ranging from $2$ to $4$. As shown in Tab.~\ref{tab:PatchSize}, MSPCaps-T achieves its best performance at $p=4$, while the larger MSPCaps-L performs optimally at $p=3$. For MSPCaps-T, a patch size of $p=3$ causes uneven division of the feature map, resulting in border feature loss and degraded accuracy. In contrast, MSPCaps-L achieves an optimal parameter balance with $p=3$, leveraging its larger capacity. However, a smaller patch size of $p=2$ introduces excessive parameterization, leading to overfitting and poor generalization. Therefore, to ensure a consistent and efficient architecture, we select $p=4$ as the global setting. This trade-off significantly improves MSPCaps-T's performance and efficiency, with only a minor accuracy reduction for MSPCaps-L.

\subsection{Adversarial Robustness Analysis}
\begin{figure}[t]
\centering
\includegraphics[width=\columnwidth]{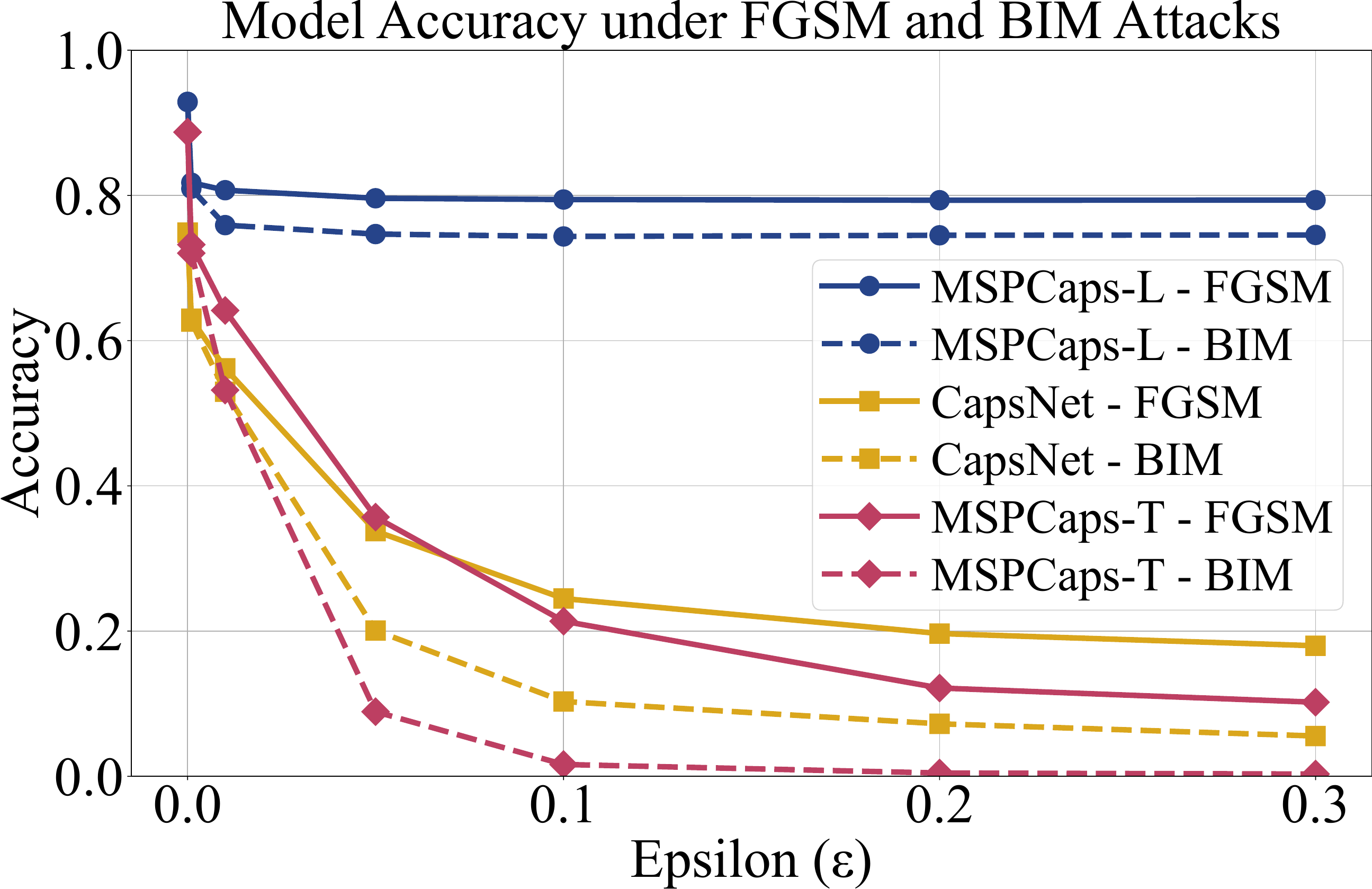}
\caption{Classification accuracy of MSPCaps-L, MSPCaps-T, and CapsNet against FGSM and BIM adversarial attacks on CIFAR-10, plotted with attack strength $\epsilon$.}
\label{Cifar10_Attack}
\end{figure}

To demonstrate the robustness of MSPCaps variants, we conduct FGSM \cite{fgsm} and BIM attacks \cite{bim} on the CIFAR-10 dataset. For the iterative attack, BIM is performed for 10 steps. The results are shown in Fig. \ref{Cifar10_Attack}. We can observe a clear distinction between our model variants. For MSPCaps-L, it exhibits exceptional stability, maintaining a high level of accuracy even as the attack strength increases. In contrast, due to parameter constraints, MSPCaps-T's performance degrades at $\epsilon>0.01$, falling below the original CapsNet's accuracy.

\subsection{Conclusion}
In this paper, we propose a novel multi-scale patchify capsule network with cross-agreement routing, named MSPCaps. We first introduce MSRB and PatchifyCaps to obtain primary capsules from multi-scale features. Then, to fuse cross-scale capsules effectively for voting, we design a CAR block to adaptively select the most coherent capsules for voting. Experimental results demonstrate that MSPCaps outperforms state-of-the-art methods in terms of classification accuracy and model robustness within capsule networks.

\section{Acknowledgments}
This work was supported in part by the Guangdong S\&T Programme under Grant No. 2024B0101030002, the Basic Research Project of Hetao Shenzhen-HK S\&T Cooperation Zone under Grant No. HZQB-KCZYZ-2021067, the National Natural Science Foundation of China under Grant No. 62501514, the Shenzhen Outstanding Talents Training Fund under Grant No. 202002, the Guangdong Research Projects under Grant No. 2017ZT07X152 and Grant No. 2019CX01X104, the Guangdong Provincial Key Laboratory of Future Networks of Intelligence under Grant No. 2022B1212010001, and the Shenzhen Key Laboratory of Big Data and Artificial Intelligence under Grant No. ZDSYS201707251409055.

\bibliography{aaai2026}

\end{document}